\pdfoutput=1

\documentclass[11pt]{article}

\usepackage{EMNLP2022}

\usepackage{times}
\usepackage{latexsym}

\usepackage[T1]{fontenc}

\usepackage[utf8]{inputenc}

\usepackage{microtype}

\usepackage{inconsolata}

\usepackage{amsmath}
\usepackage{graphicx}
\usepackage{multirow}
\usepackage{booktabs}
\usepackage{amsfonts}
\usepackage[ruled,linesnumbered]{algorithm2e}

\title{Capturing Global Structural Information in Long Document Question Answering with Compressive Graph Selector Network}

\author{Yuxiang Nie${}^{123}$,~~Heyan Huang${}^{123}$\thanks{~~Corresponding author},~~Wei Wei${}^{4}$,~~Xian-Ling Mao${}^{123}$ \\
        ${}^{1}$School of Computer Science and Technology, Beijing Institute of Technology\\
        ${}^{2}$Beijing Engineering Research Center of High Volume Language Information Processing \\
and Cloud Computing Applications\\
${}^{3}$Beijing Institute of Technology Southeast Academy of Information Technology\\
${}^{4}$Huazhong University of Science and Technology\\
        \texttt{\{nieyx,hhy63,maoxl\}@bit.edu.cn,weiw@hust.edu.cn}}

\begin{document}
\maketitle
\begin{abstract}

Long document question answering is a challenging task due to its demands for complex reasoning over long text. Previous works usually take long documents as non-structured flat texts or only consider the local structure in long documents. However, these methods usually ignore the global structure of the long document, which is essential for long-range understanding. To tackle this problem, we propose \textbf{C}ompressive \textbf{G}raph \textbf{S}elector \textbf{N}etwork (CGSN) to capture the global structure in a compressive and iterative manner. The proposed model mainly focuses on the evidence selection phase of long document question answering. Specifically, it consists of three modules: local graph network, global graph network and evidence memory network. Firstly, the local graph network builds the graph structure of the chunked segment in token, sentence, paragraph and segment levels to capture the short-term dependency of the text. Secondly, the global graph network selectively receives the information of each level from the local graph, compresses them into the global graph nodes and applies graph attention to the global graph nodes to build the long-range reasoning over the entire text in an iterative way. Thirdly, the evidence memory network is designed to alleviate the redundancy problem in the evidence selection by saving the selected result in the previous steps. Extensive experiments show that the proposed model outperforms previous methods on two datasets.\footnote{We have released our codes and data in \url{https://github.com/JerrryNie/CGSN}.}

\end{abstract}

\section{Introduction}

Long document question answering (LDQA)  is a task to select relevant evidence and answer questions over long text \citep{dasigi2021dataset}. Compared to the traditional QA tasks, whose input is often under 512 tokens\footnote{In this paper, `token' means sub-tokens split from a text sequence by a specific pre-trained tokenizer.}, the input of LDQA can be more than 20K tokens.

LDQA methods can be divided into two categories: end-to-end methods and select-then-read methods. 
The end-to-end methods usually take a question and a long text as input to select evidence and produce the answer in one step.  For example, \citet{dasigi2021dataset} use Longformer-Encoder-Decoder (LED) model to select evidence in the encoder part and generate answers in the decoder part. The select-then-read methods firstly apply an evidence selection model to obtain evidence pieces in a long document and then use an answer generation model to generate answers given the evidence pieces and the question. These methods mainly focus on the evidence selection phase. For example, \citet{karpukhin2020dense} and \citet{zhu2021adaptive} select paragraphs in an open domain retrieving manner. \citet{zheng2020document} and \citet{ainslie2020etc} build structure on the chunked documents for evidence selection. \citet{gong2020recurrent} model information flows among chunks to enhance the ability of the model in selecting the evidence. However, most of the two kinds of works ignore the global structure of a long document when selecting evidence pieces, which is crucial to long-range understanding.
Therefore, improvement on the evidence selection phase is needed.

Motivated by the human reading process: selectively memorizing the important pieces of information and integrating them, we propose an evidence selection model in the select-then-read method, named \textbf{C}ompressive \textbf{G}raph \textbf{S}elector \textbf{N}etwork (CGSN). It aims to capture the global structural information in a compressive and iterative manner. Specifically, the model is composed of three modules: the local graph network, the global graph network and the evidence memory network. Firstly, the local graph takes a segment\footnote{A `segment' is a series of paragraphs in a document.} of the document as input and implements graph attention among tokens, sentences, paragraphs and the segment itself. Secondly, the global graph module selectively receives the information from the local graph and compresses it with the stored information via multi-head attention. Then, the graph attention is applied to the global graph to integrate the global structural information, which is written back to the local graph nodes to enhance the expression of local nodes for evidence selection. Thirdly, the evidence memory network receives and summarizes the evidence selecting results and sends them into the global network to alleviate the evidence redundancy problem.

Extensive experiments on two datasets show that CGSN outperforms previous methods in the evidence selection phase. Using the same answer generator as the previous methods do, CGSN further reaches the best results in the answer generation phase.

Our contributions are as follows:
\vspace{-5pt}
\begin{itemize}
\setlength{\itemsep}{1pt}
\setlength{\parsep}{0pt}
\setlength{\parskip}{0pt}
    \item To the best of our knowledge, we are the first to consider the global structure in the long document QA task.
    \item With the enhancement of global structural information, the proposed model, CGSN outperforms previous methods.
\end{itemize}

\section{Related Works}
\paragraph{Long Document Question Answering.} Long Document question answering aims to answer the question with the comprehension of the long document and applies multi-hop reasoning among retrieved evidence paragraphs. \citet{dasigi2021dataset} take advantage of the pre-trained model LED \citep{beltagy2020longformer} and treat the input as a long sequence to predict the evidence paragraphs and generate the answer. \citet{zheng2020document} and \citet{ainslie2020etc} model the structure on the chunked document to select the evidence paragraph. Although \citet{ainslie2020etc} claims that they explicitly model the structure of long documents, the input of their model is limited in 4K tokens, which can be regarded as a relatively long chunk. \citet{gong2020recurrent} use the recurrent mechanism to enable information flow through different chunks for evidence selection. \citet{karpukhin2020dense} and \citet{zhu2021adaptive} search for relevant evidence from individual paragraphs in the long document. However, most of these works model the long document as a flat sequence or consider the local structure in the document segments while global structural information of the document is nearly neglected.

\paragraph{Graph Neural Networks.} Graph neural network (GNN) is popular in various tasks \citep{yao2019graph,schlemper2019attention} due to its effectiveness in modeling structural information. Among different variants of GNNs, Graph Attention Network \citep{velivckovic2017graph} (GAT) can take advantage of the attention mechanism in a graph, attending neighborhood node features to the node by different attention weights. \citet{zheng2020gman} make use of a graph multi-attention network to predict traffic conditions. \citet{abu2018watch} take advantage of the graph attention to automatically guide the random walk in graph generation. In natural language tasks, due to the limit of memory usage, GAT is often used to model short sequences. Therefore, modeling the graph structure of the long sequence is nearly unexplored.
\begin{figure*}
\centering
\includegraphics[width=.9\textwidth]{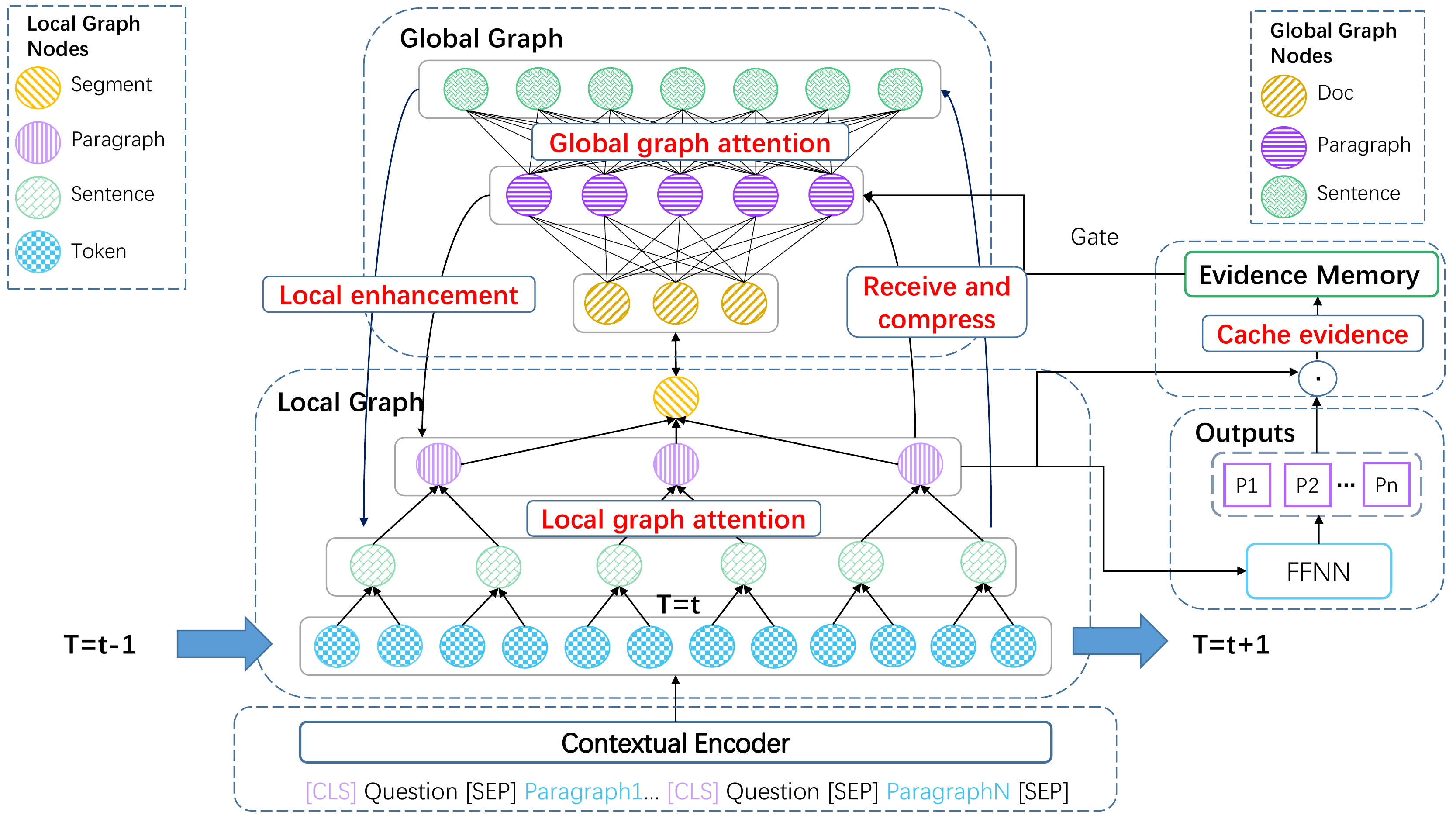}
\caption{\label{fig:model}The architecture of CGSN.}
\end{figure*}

\paragraph{Memory Networks.} Memory network \citep{Weston2015MemoryN} is used in memorizing long-term information via learnable reading/writing components. It is first applied to the QA task for knowledge base reasoning, which also achieves much progress in summarization \citep{cui2021sliding} and visual question answering. To memorize plenty of information, the memory network learns to read and recurrently write into an external memory via attention. \citet{miller2016key} propose Key-Value Memory Network to flexibly access knowledge for question answering. \citet{lu2020chime} design a context memory for cross-passage evidence reasoning. However, these methods only consider the memory on a single level, while structural information is disregarded.

\section{Compressive Graph Selector Network}
In this section, we first formalize the long document question answering (LDQA) task, and then introduce the proposed  evidence selection model, i.e. \textbf{C}ompressive \textbf{G}raph \textbf{S}elector \textbf{N}etwork (CGSN). As for the answer generator, we use a vanilla LED as the answer generator and describe the implementation details in Appendix \ref{appendix:answer_generator}. Finally, we discuss the advantages of the select-then-read methods over the end-to-end methods.

\subsection{Problem Formulation}
The input to LDQA is a question $\boldsymbol{q}=[q_1,q_2,...,$ $ q_m]$ coupled with a document $\boldsymbol{d} = [\boldsymbol{p}_1,\boldsymbol{p}_2,...,$ $\boldsymbol{p}_n], \boldsymbol{p}_i = [t_{i1}, t_{i2}, ..., t_{i, k_i}] (1\leq i \leq n)$ where $m$ denotes the length of the question, $n$ denotes the number of paragraphs in the document and $k_i$ denotes the length of paragraph $i$. The length of the document is defined as the sum of the length of each paragraph: $c=\sum_{i}^{n}k_i$. In the LDQA setting, the length $c$ is often unlimited, which can be larger than 20K. The goal of LDQA is to produce the evidence paragraphs $\{\boldsymbol{p}_{e_i}\}_{i=1}^{e^q}$ and generate the free-form answer $a=[a_1,a_2,...,a_r]$ based on $\boldsymbol{q}$ and $\boldsymbol{d}$, where $\boldsymbol{p}_{e_i}$ denotes the $e_i$th paragraph in the document (the $i$th paragraph in the evidence set), $e^q$ is the number of evidence for question $\boldsymbol{q}$.

\subsection{Overview of the Model}
To explore the global graph structure of long sequences, we propose Compressive Graph Selector Network (CGSN), which operates in an iterative and compressive way. CGSN is composed of three modules, local graph network, global graph network and evidence memory network. As shown in Figure~\ref{fig:model}, firstly, in time step\footnote{The `time step' is the order of the segment to be processed.} $T=t$, the local graph network takes the $t\text{th}$ segment of a document as the input and models the graph structure in token, sentence, paragraph and segment levels. Secondly, the global graph selectively receives the information from each granularity, compresses them into the corresponding global graph nodes, implements the graph attention among the global graph nodes and sends the global-attended information back to the local graph to enhance the expression of the local graph nodes for evidence selection. Thirdly, the evidence memory network receives the enhanced paragraph nodes from the local graph, summarizes and caches them via the predicting logits. At the beginning of the time step $T=t+1$, the stored memory is sent and  fused with the global graph nodes in order to alleviate the redundant evidence selection problem. The detailed architecture is described in Appendix~\ref{sec:algorithm}.

\subsection{Local Graph Network}
\label{sec:local_graph_network}

\paragraph{Input Format} Let $\textbf{Seg}_k=[\boldsymbol{p}_{k,1},...,\boldsymbol{p}_{k,N_{\text{seg}}}]$ be the $k$th segment in a document, which composed of $N_{\text{seg}}$ paragraphs. To build the local graph network, firstly, we encode the $N_{\text{seg}}$ paragraphs paired with the question. For each question-paragraph pair, the input format is ``\texttt{[CLS] q [SEP] $\text{p}_\text{i}$ [SEP]}'', where $1\leq i \leq N_{\text{seg}}$. We set the embeddings of each input pair as $\mathbf{E}\in\mathbb{R}^{\ell\times d_w}$, where $\ell$ is the length of the input and $d_w$ is the dimension of the embedding. The $N_{\text{seg}}$ embedding sequences are stacked as $\mathbf{E}_k\in\mathbb{R}^{N_{\text{seg}}\times \ell\times d_w}$ and sent into the encoder $f_e$ as follows:
\noindent
\begin{equation}
    \mathbf{H}_k = f_e(\mathbf{E}_k)
\end{equation}
\noindent
where $\mathbf{H}_k\in\mathbb{R}^{N_{\text{seg}}\times \ell\times d_h}$ is the contextual encoding, $d_h$ denotes its dimension. In general, we use pre-trained Transformer encoders \citep{devlin2019bert,beltagy2019scibert}.

\paragraph{Local Graph Initialization} Let $\mathbf{h}^{L}_{t_i}$, $\mathbf{h}^{L}_{\text{sent}_i}$, $\mathbf{h}^{L}_{p_i}$, $\mathbf{h}^{L}_\text{seg}$ be the node representations of the local graph in token, sentence, paragraph and segment level, where $t_i$ denotes the $i$th token in the segment, $\text{sent}_i$ denotes the $i\text{th}$ sentence, $p_i$ denotes the $i\text{th}$ paragraph. And the question-aware token hidden states of the segment  $\mathbf{H}_k^{\text{Seg}}\in\mathbb{R}^{\ell_k\times d_h}$ are used to initialize $\mathbf{h}^{L}_{t_i}(1\leq i \leq \ell_k)$, where $\ell_k$ is the number of tokens in a segment. The sentence node $\mathbf{h}^{L}_{\text{sent}_i}$ is initialized by the mean-pooling of $\mathbf{h}^{L}_{t_i} (i\in\mathcal{S}_i)$, where $\mathcal{S}_i$ denotes the set of token indexes of sentence $i$. The paragraph node $\mathbf{h}^{L}_{p_i}$ is initialized by the encodings of ``\texttt{[CLS]}'' token and the segment node $\mathbf{h}^{L}_\text{seg}$ is initialized by the mean-pooling of all the paragraph nodes.\footnote{We also experiment CGSN with ``LED Encoder''. The input is ``\texttt{<s> q </s> </s> $\text{p}_\text{i}$ </s>}''. We use the last ``\texttt{</s>}'' token to initialize the paragraph node in the local graph.}

To enhance the interaction among the nodes at the same level, we apply the BiLSTM in the token-level, sentence-level and paragraph-level nodes, and fuse the encoded states with the node representation by a feed-forward neural network (FFNN) to generate the ultimate node representation:
\noindent
\vspace{-1pt}
\begin{gather}
    \mathbf{h}^{L}_{\text{LSTM}}=\text{BiLSTM}(\mathbf{h}^{L})\\
    \mathbf{h}^L_\text{init}=\text{FFNN}(\mathbf{h}^{L}_{\text{LSTM}},\mathbf{h}^{L})    
\end{gather}
\noindent
where $\mathbf{h}^L=[\mathbf{h}^L_1,...,\mathbf{h}^L_m]$ denotes the set of nodes at the same level ordered by the text sequence.

\paragraph{Local Graph Attention} After initialization, the multi-hop graph attention is applied among the local graph nodes. The aim is to capture the local structural information for understanding the current segment of paragraphs. Let the edge set of the local graph $\mathcal{G}^L$ be $\mathcal{E}=\{(i,j)\}_{(i,j)\in \mathcal{G}^L}$, where the $(i,j)$ denotes an edge from node $i$ to node $j$. In the proposed model, the edges of the local graph are predefined as unidirectional, which can only send information from the lower level into the neighboring higher one, including token to sentence, sentence to paragraph and paragraph to the segment node. The hidden output of the graph attention is then fused with the graph node representation and then used in the graph attention operation in the next hop. Let $\mathbf{h}_j^o$ be a node representation in the local graph of hop $o$, followed \citet{velivckovic2017graph}, its representation in hop $o+1$ is:
\noindent
\begin{equation}
    \text{e}_{ij}=\frac{(\mathbf{h}_j^o\mathbf{W}^Q)(\mathbf{h}_i^o\mathbf{W}^K)^\text{T}}{\sqrt{d_z}}
\end{equation}
\noindent
where $e_{ij}$ denotes the attention coefficients from node $i$ to node $j$. To normalize them, there is:
\begin{equation}
    \alpha_{ij}=\text{softmax}_i(e_{ij})=\frac{\text{exp}(e_{ij})}{\sum_{k\in\mathcal{N}_j}\text{exp}(e_{kj})}
\end{equation}
Therefore, the output of the attention operation is:
\noindent
\begin{gather}
    \mathbf{z}_j^{\text{head}_m}=\sum_{i\in\mathcal{N}_j}\alpha_{ij}\mathbf{h}_i\mathbf{W}^V\\
    \mathbf{h}_j^{o+1}=\text{Cat}[\mathbf{z}_j^{\text{head}_1},...,\mathbf{z}_j^{\text{head}_k}]
\end{gather}
\noindent
where $\mathbf{W}^Q,\mathbf{W}^K,\mathbf{W}^V$ are the parameters of the attention query, key and value of $\text{head}_m$, $\mathcal{N}_j$ is the neighboring nodes of node $j$, $\mathbf{z}_j^{\text{head}_m}$ is the output of $\text{head}_m$ in the multi-head attention, and all these outputs are finally concatenated into the vector $\mathbf{h}_j^{o+1}$, which serves as the representation of node $j$ in the time step $o+1$.

\subsection{Global Graph Network}

The global graph is composed of three types of nodes: global sentence nodes $\mathbf{h}_{\text{sent},i}^G\ (1\leq i \leq N^G_{\text{sent}})$, global paragraph nodes $\mathbf{h}_{\text{p},i}^G\ (1\leq i \leq N^G_{\text{p}})$ and global document nodes $\mathbf{h}_{\text{d},i}^G\ (1\leq i \leq N^G_{\text{d}})$, where the token-level nodes are neglected because it's hard to memorize a large number of the tokens in the finite nodes. $N^G_{\text{sent}},N^G_{\text{p}}$ and $N^G_{\text{d}}$ are predefined node numbers in each level.

\paragraph{Local Information Reception and Compression}

Global structural reasoning is based on the interaction among local information. Therefore, to store and process local information efficiently, each level of global graph nodes only receives information from one level in the local graph. The sentence-level global graph nodes receive the information from the sentence-level local nodes, and the same thing happens between the two kinds of paragraph nodes. In particular, the document-level nodes in the global graph receive information from the segment-level nodes in the local graph. All of the receptions are done with the multi-head attention mechanism. This procedure is similar to that in Section~\ref{sec:local_graph_network}. The only difference is that the direction of the attention is from all of the local graph nodes to the global graph node at the corresponding level. For each global graph node i, let $\mathbf{h}^G_{i,\text{local}}$ be its local-aware global graph node representation. Based on the information from the local graph nodes, a gated network is used to fuse the representations of the global graph node $\mathbf{h}_i^G$ and the local-aware global node $\mathbf{h}^G_{i,\text{local}}$:

\noindent
\begin{gather}
    \mathbf{z}^{G}_{i,\text{fuse}}=\text{FFNN}(\mathbf{h}_i^G,\mathbf{h}^G_{i,\text{local}})\\
    \gamma=\text{Gate}(\mathbf{z}^{G}_{i,\text{fuse}})\\
\vspace{-10pt}
    \mathbf{h}_{i,\text{fuse}}^G=(1-\gamma)\mathbf{h}_i^G+\gamma \mathbf{z}^{G}_{i,\text{fuse}}
\end{gather}
\noindent
where $\mathbf{h}^{G}_{i,\text{fuse}}$ is the representation of the updated global graph node $i$.

\paragraph{Global Graph Attention}
To accomplish the global structural reasoning process, we implement the \textit{global} graph attention. As shown in Figure~\ref{fig:model}, for each pair of nodes between different levels of the Global Graph, cross attention is used to extract information from one level of nodes to the other, including sentence-paragraph, paragraph-document and document-sentence (not displayed). This process is executed $m$ times to fully extract global structural information. The applied attention is similar to Section~\ref{sec:local_graph_network}.

\paragraph{Enhance the Local Nodes with Global Representation}

To enhance the local expression, the local paragraph nodes attend information from the global paragraph nodes by multi-head attention and combine the attended representations with the local paragraph nodes via a feed-forward network. Let the global-enhanced local representation of paragraph $j$ be $\mathbf{h}^{L\leftarrow G}_j$. The evidence selection loss is:
\noindent
\begin{equation}
    \mathcal{L}=-\frac{1}{N}\sum_{i=1}^N\sum_{j=1}^{n_i}[\text{log}\  P(b|\mathbf{h}^{L\leftarrow G}_j)]
\end{equation}
\noindent
where $N$ is the number of training instances, $n_i$ is the number of paragraphs in the instance $i$ and $b$ is the binary label (``1'' denotes the paragraph is an evidence paragraph, ``0'' otherwise).

\begin{table}[t]
\fontsize{9.5}{10}\selectfont
    \centering
    \begin{tabular}{lcccc}    
        \toprule
        \multirow{2}{*}{Models} & \multicolumn{3}{c}{Answer F1}\\
        & Extractive & Abstractive & Overall\\
        \midrule
        LED (doc) & 28.22 & 14.00 & 23.58\\
        LED (gold) & 52.41 & 23.44 & 42.97   \\
        \bottomrule   
    \end{tabular}
    \caption{\label{tab:gold_evidence_comparison} The answer F1 scores on the test set of Qasper. ``doc'' and ``gold'' denote the input types: the entire long document/gold evidence.} 
\end{table}

\subsection{Evidence Memory Network}

The evidence memory network aims to integrate the evidence-selecting decision and enhance the results into the global network for the next time step. It is composed of a cache and a writing network. To properly update the parameters of the network, all of the manipulations are done in the next time step, where we assume that the current time step is $t$ and the next time step is $t+1$. In the time step $t$, the evidence predicting logits of time $t$ are converted via the softmax function into weights, which are multiplied with the corresponding local paragraph nodes, summed and cached in the time $t$:
\noindent
\begin{gather}
    \alpha_i=\frac{\text{exp}(e_i)}{\sum_{j=1}^n\text{exp}(e_j)}\\
    \mathbf{h}^L_{\text{sum}}=\sum_{i=1}^n\alpha_i\mathbf{h}^{L\leftarrow G}_i
\end{gather}
\noindent
where $e_i$ is the logit of paragraph $i$ in the evidence selection step, $n$ is the number of paragraphs and $\mathbf{h}^{L\leftarrow G}_i$ is the global-enhanced representation of paragraph node $i$ in the local graph.

At time $t+1$, $\mathbf{h}^L_{\text{sum}}$ is fused with the global graph paragraph nodes $\mathbf{h}^G_{\text{p},i}$ via the writing network:
\noindent
\begin{gather}
    \mathbf{h}^G_{i,\text{cat}} = \text{C}\text{at}[\mathbf{h}^G_{\text{p},i},\mathbf{h}^L_{\text{sum}}]\\
    \mathbf{h}^G_{i,\text{merge}} = \text{F}\text{FNN}(\mathbf{h}^G_{i,\text{cat}})\\
    \gamma=\text{Gat}\text{e}(\mathbf{h}^G_{i,\text{cat}})\\
    \mathbf{h}^G_{i,\text{updated}} = (1-\gamma)\mathbf{h}^G_{\text{p},i}+\gamma\mathbf{h}^G_{i,\text{merge}}
\end{gather}
\noindent
where $\mathbf{h}^G_{i,\text{updated}}$ is the updated paragraph node $i$ of the global graph in time $t+1$.

\subsection{Select-then-read Methods versus End-to-end Methods}

Intuitively, end-to-end methods can learn to produce answers and usually regard evidence selection as an auxiliary task. However, there are two disadvantages to these methods. Firstly, evidence selection is essential. According to Table~\ref{tab:gold_evidence_comparison}, given the gold evidence, the performance of the LED on answer generation improves by a large margin. 
Secondly, the long input is memory-consuming. In the long input setting, with sparse attention models \citep{beltagy2020longformer}, due to the relatively large local window size, the demands for GPU memory still become high. Meanwhile, the average token length of gold evidence in Qasper \citep{dasigi2021dataset} is only 149.81, which shows that it's probably memory-efficient only to regard the selected evidence as the context to generate answers. Therefore, a well-designed evidence selection model in the select-then-read method can be beneficial.

\section{Experimental Setup}

\subsection{Datasets}
\begin{figure}
\centering
\includegraphics[width=.8\linewidth]{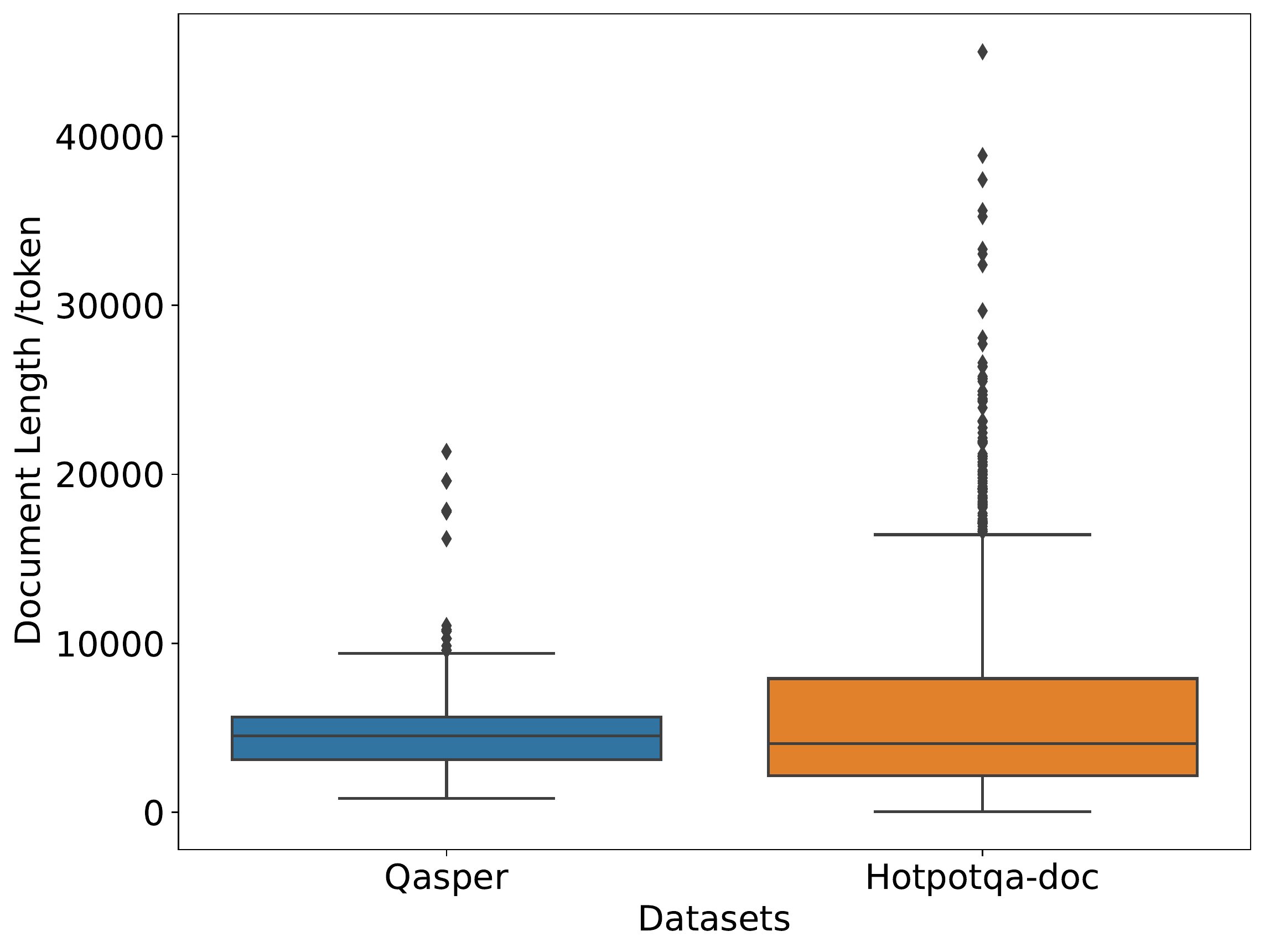}
\caption{\label{fig:stat-of-doc-len-distribution}The box plots of document length distribution on Qasper and HotpotQA-Doc.}
\end{figure}

\paragraph{Qasper} Qasper \citep{dasigi2021dataset} is a QA dataset for long scientific papers. It aims to select the evidence and make multi-hop reasoning to produce answers given the long text and the question. It contains extractive, abstractive, yes/no and unanswerable answers. Followed \citet{dasigi2021dataset}, we treat all the answers as free-form answers. 

\begin{table*}[t]
\fontsize{9.7}{10}\selectfont
    \centering
    \begin{tabular}{lcccc|cc}    
        \toprule
        Models& Extractive & Abstractive & Yes/No & Unanswerable & Evidence & Overall\\
        \midrule
        LED \citep{dasigi2021dataset}*  & 28.22 & 14.00 & 65.88 & 51.92 & 29.20 & 33.12\\
        ETC \citep{ainslie2020etc} & 32.88 & 13.49 & 58.00 & 62.28 & 51.17 & 35.37\\
        DPR \citep{karpukhin2020dense} & 22.69 & 9.73 & 56.10 & 69.02 & 36.98 & 30.22\\
        AISO \citep{zhu2021adaptive} & 26.66 & 10.79 & 55.56 & 70.72 & 42.74 & 32.52\\
        RCM \citep{gong2020recurrent} & 30.69 & 11.95 & 64.32 & 70.39 & 47.28 & 35.99\\
        BERT-DM \citep{zheng2020document} & 31.84 & 14.38 & 63.86 & 66.86 & 48.09 & 36.52\\
        \midrule
        AISO (SciBERT) & 28.86 & 11.90 & 60.30 & 71.51 & 45.92 & 34.30\\
        RCM (SciBERT) & 30.85 & 13.92 & 63.45 & 68.53 & 47.69 & 35.88\\
        BERT-DM (SciBERT) & 32.30 & 12.98 & 62.56 & 71.69 & 48.52 & 36.63\\
        \midrule        
        CGSN (LED Encoder) & 31.11 & 12.52 & 63.00 & \textbf{77.90} & 49.55 & 37.21\\
        CGSN (SciBERT) & \textbf{34.75} & \textbf{14.39} & \textbf{68.14} & 71.84 & \textbf{53.98} & \textbf{39.44}\\
        \midrule
        LED (Gold Evidence) & 52.41 & 23.44 & 76.96 & 77.91 & - & 52.87\\
        Human & 58.92 & 39.71 & 78.98 & 69.44 & 71.62 & 60.92\\
        \bottomrule   
    \end{tabular}
    \caption{\label{tab:main_result_qasper} The performance of F1 scores on the test set of Qasper. In the first row, ``Extractive, Abstractive, Yes/No, Unanswerable'' are four types of answers. ``Evidence'' is the F1 score of evidence selection. ``Overall'' is the F1 score of all the answers. ``*'' denotes a reimplementation, which is slightly different from the reported result in the original paper. ``SciBERT'' and ``LED Encoder'' are backbone models for evidence selection methods in some of our implementations. Other implementations use the backbone models in their original papers. The last two rows are the upperbounds of the task, where the gold evidence is used or the human expert is engaged.} 
\end{table*}

\paragraph{HotpotQA-Doc} HotpotQA \citep{yang2018hotpotqa} is a dataset to answer multi-hop questions based on the supporting fact sentences. 

We create a long document version of the HotpotQA dataset, called \textit{HotpotQA-Doc}. To build the dataset, we select all the hard-level questions where multi-hop reasoning is required, crawl the corresponding Wikipedia\footnote{\url{https://www.wikipedia.org}} articles containing these sentences and concatenate these articles as a long document. Each paragraph with the supporting fact sentence is regarded as a gold evidence paragraph. We remove the QA pairs whose supporting facts cannot be found in any article. We split the original dev set into a new dev set and test set. We finally collect 15,658/2,468/4,937 QA pairs in train/dev/test set. Due to the change of the content in the Wikipedia pages, some answers are not guaranteed to be the span in the content but are still similar. Therefore, we regard all the answers as free-form answers.

The boxplot of Qasper and HotpotQA-Doc are shown in Figure~\ref{fig:stat-of-doc-len-distribution}.

\subsection{Implementation Details} 
\label{sec:experimental_setup}
We experiment on two backbones: the  SCIBERT-SCIVOCAB-uncased version of SciBERT \cite{beltagy2019scibert} and the encoder of LED-base-16384 \cite{beltagy2020longformer}. In the SciBERT setting, the local hop number is 4 while in the LED Encoder, the local hop number is 2. For both the two backbones, global hop is 1, paragraph number in a segment is 16, global graph sentence nodes are 64, paragraph nodes are 32, and document nodes are 4. We searched over \{1e-6, 5e-6, 1e-5, 5e-5\} learning rates, \{5\%, 10\%, 20\%, 30\%\} warmup proportions, \{1, 0.1, 0.01, 0.001\} weight decays, \{2, 4, 8\} batch sizes and \{1, 3, 5, 7\} epochs. And the final learning rate is 1e-5, the warmup proportion is 10\%, the weight decay is 0.01, the batch size is 4 and the epoch number is 5. We train an LED model (called \textit{LED (evidence-trained)}) with the QA pairs and gold evidence in the training set as the answer generator (Appendix \ref{appendix:answer_generator}). It receives selected evidence pieces in the testing phase from evidence-selecting methods (including CGSN and other evidence selection baselines) to further evaluate the performance of these evidence-selection methods on answer generation. It took 3.33 hours to train the model on four 11GB GPUs.

\begin{table}[t]
\fontsize{9.5}{10}\selectfont
    \centering
    \begin{tabular}{lcc}    
        \toprule
        {Models} & E-F1 & A-F1\\
        \midrule
        LED \cite{dasigi2021dataset} & 68.36 & 51.50 \\
        LED Encoder \cite{beltagy2020longformer} & 76.02 & 53.99 \\
        ETC \cite{ainslie2020etc} & 91.01 & 57.01 \\
        DPR \cite{karpukhin2020dense} & 87.81 & 56.62 \\
        AISO \cite{zhu2021adaptive}& 88.28 & 55.61 \\
        RCM \cite{gong2020recurrent}& 90.03 & 57.20 \\
        BERT-DM \cite{zheng2020document}& 91.45 & 57.63 \\
        \midrule
        AISO (BERT) & 88.97 & 56.04 \\
        \midrule
        CGSN (LED Encoder) & \textbf{92.02} & 57.80 \\
        CGSN (SciBERT) & 92.02 & 57.79 \\
        CGSN (BERT) & 91.88 & \textbf{57.92} \\
        \midrule
        LED (Gold Evidence) & - & 58.94\\
        \bottomrule   
    \end{tabular}
    \caption{\label{tab:main_result_hotpotqa_doc} The performance of Evidence F1 (E-F1) and Answer F1 (A-F1) on the test set of HotpotQA-Doc. ``BERT'' denotes the model ``BERT-base-uncased''.} 
\end{table}

\section{Experiments}
In this section, firstly, we discuss the main result of the CGSN model on Qasper and HotpotQA-Doc, and then further explore the proposed model.
\subsection{Main Results}

\begin{table*}[t] 
    \centering
    \begin{tabular}{lcccc}    
        \hline
        \multirow{2}{*}{Models} & \multicolumn{2}{c}{SciBERT}& \multicolumn{2}{c}{LED Encoder}\\
        & Evidence F1 & Answer F1 & Evidence F1 & Answer F1\\
        \hline
        CGSN & \textbf{49.72}& \textbf{33.20}& \textbf{41.55}& \textbf{29.76}\\
        \hline
        w/o Local Graph & 47.86& 32.80& 38.85& 28.33\\
        w/o Global Edges & 48.13& 31.14&39.55& 27.86\\
        w/o Global Graph & 48.09& 31.70& 39.54& 28.10\\
        w/o Evidence Memory & 49.21& 32.61& 40.78& 29.30\\
        \hline   
    \end{tabular}
    \caption{\label{tab:ablation} Ablation study of CGSN with SciBERT/LED Encoder, evaluating on the dev set of Qasper.} 
\end{table*}

We evaluate the performance of the end-to-end LED model \cite{dasigi2021dataset} and the evidence selection methods\footnote{These evidence selection methods are coupled with LED(evidence-trained) to evaluate the performance on answer generation.}, including the proposed CGSN, the encoder of LED model\footnote{We only report its performance on HotpotQA-Doc because it is hard to converge on the training set of Qasper.}, the open domain retrieving method DPR \cite{karpukhin2020dense} and AISO \cite{zhu2021adaptive}, evidence selection part of RCM \cite{gong2020recurrent}, BERT-DM \cite{zheng2020document} and ETC \citep{ainslie2020etc}. We use evidence F1 and answer F1 \citep{dasigi2021dataset} as the metrics. All the retrieving methods only retrieve candidate paragraphs from the specific document. For AISO, we remove the link retrieval and stop evidence selection when the answer action is taken. In RCM, BERT-DM and ETC, we keep the evidence selection loss. 

The main result of the CGSN on Qasper is shown in Table~\ref{tab:main_result_qasper}. It can be found that CGSN (SciBERT) outperforms all the baseline methods in evidence selection and answer generation. Although the pre-trained model ETC can deal with relatively long sequences (4K) and construct the structure via global-local attention, CGSN still outperforms it. For a fair comparison, we also use SciBERT as the backbone in AISO, RCM and BERT-DM. The result indicates the advantages of CGSN. Besides, to fairly compare CGSN with the LED, we also take the encoder of LED as the backbone of CGSN. Although LED Encoder is not designed as a discriminative model, CGSN (LED Encoder) also obtains strong performance.

Additionally, we also evaluate the proposed model on the HotpotQA-Doc dataset. As shown in Table~\ref{tab:main_result_hotpotqa_doc}, we notice that in evidence F1, most of the methods can reach high scores. This is because, in HotpotQA, sentence-level understanding is enough to select evidence without considering the entire Wikipedia article. Therefore, the method without long-range modeling can get a good performance. However, the result shows that with the enhancement of the global graph, the proposed CGSN can further improve its performance.

\subsection{Ablation Study}
We conduct an ablation study on components of the CGSN model. As shown in Table~\ref{tab:ablation}, firstly, we can see the performance drops when removing the local graph, which suggests its value on segment-level reasoning. Secondly, when the global graph edges are ablated, the performance of the model is lower. Furthermore, when removing the global graph, the performance of the model is even lower, showing the importance of the global graph. Finally, the result in the last row shows that evidence memory also plays a part in the CGSN model.
\begin{table}[t]
  \centering
  \begin{tabular}{cccc|cc}
    \toprule
    \multicolumn{4}{c}{Combinations} & \multicolumn{2}{c}{SciBERT}\\
    \midrule
    $N_{\text{seg}}$& $N^G_{\text{sent}}$& $N_{\text{p}}^G$& $N^G_{\text{d}}$& E-F1 & A-F1\\
    \midrule
    16& 64 & 32 & 4 & \textbf{49.72} & \textbf{33.20}\\
    8& 64 & 32 & 4 & 48.33 & 32.85\\
    16& 48 & 32 & 4 & 48.18 & 31.95\\
    16& 64 & 16 & 4 & 48.66 & 31.72\\
    16& 64 & 32 & 2 & 48.57 & 33.01\\
    16& 64 & 80 & 4 & 48.86 & 32.13\\
    16& 64 & 32 & 48 & 49.08 & 32.45\\
    \bottomrule
  \end{tabular}
  \caption{The Evidence F1 (E-F1)/Answer F1 (A-F1) under different combinations of paragraph number in a segment ($N_{\text{seg}}$), the node number in sentence ($N_{\text{sent}}^G$), paragraph ($N_p^G$) and document ($N_d^G$) levels of the global graph, evaluated on the dev set of Qasper.}
  \label{tab:chunk-combination-experiment}
\end{table}

\subsection{Effects of Paragraph Number and Global Node Number}
We conduct the experiment to explore the effectiveness of the paragraph number in a segment as well as the global graph nodes at each level. As shown in Table~\ref{tab:chunk-combination-experiment}, firstly, compared between line 1 and line 2, it can be found that with the increase of paragraph number in a segment, the performance of the model gains. It is intuitive because when the number of paragraphs increases, the ``perception field'' of the local graph becomes larger, which leads to better performance. Secondly, compared to lines 1,3,4 and 5, it can be observed that line 1 is a proper setting for the global graph. Thirdly, compared among lines 1,6 and 7, it suggests that it is proper to set the number of sentence nodes larger than that of paragraph nodes, and the same relation maintains between paragraph nodes and document nodes.

\subsection{Effects of Graph Hop Number}
\begin{figure}[t]
\centering
\includegraphics[width=\linewidth]{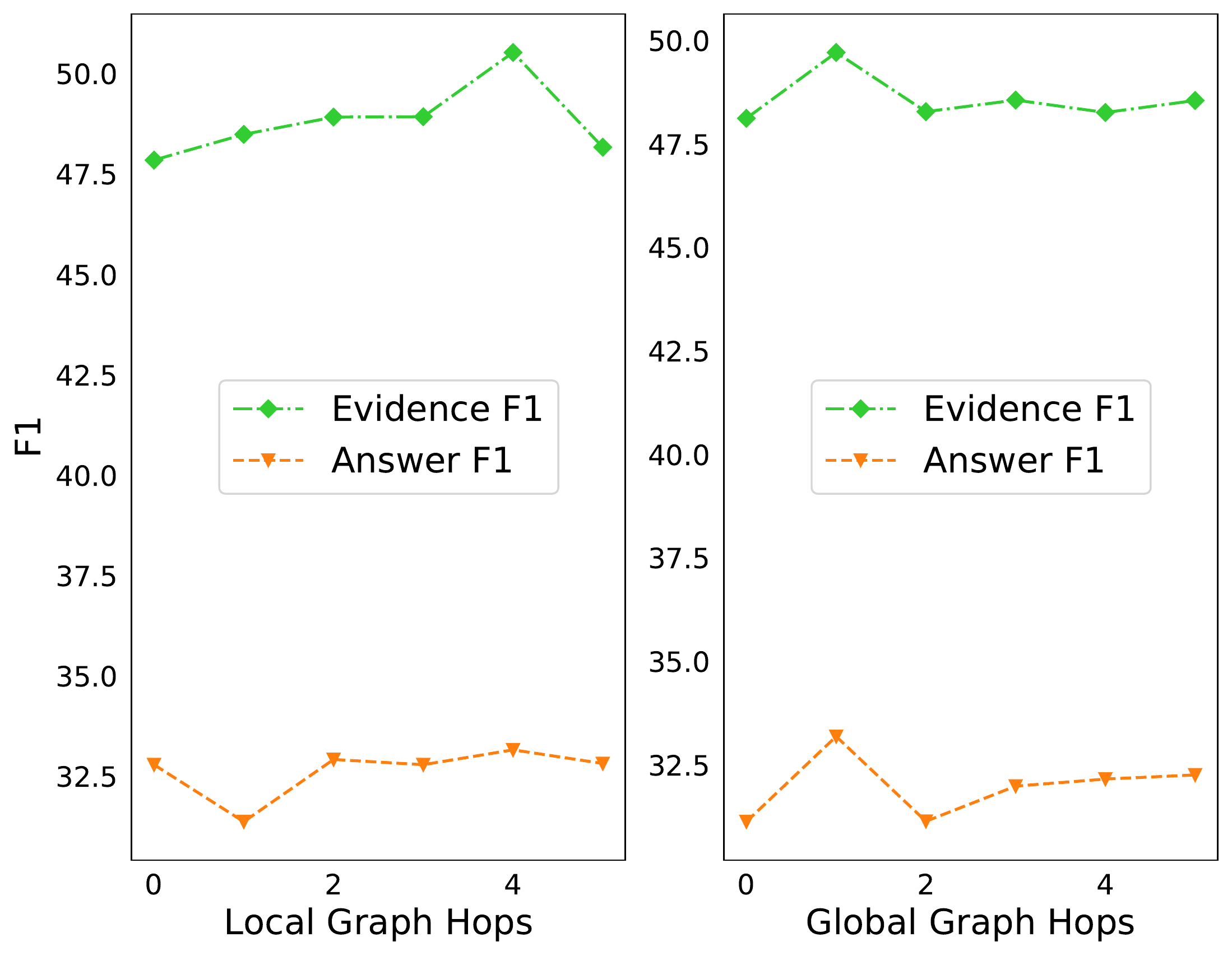}
\caption{\label{fig:local-and-global-gat-hops-qasper}Evidence/Answer F1 scores of the proposed CGSN model with varied local/global GAT hops, evaluated on the dev set of Qasper.}
\end{figure}

We conduct the experiment on the number of attention hops on the local graph and on the global graph. The result is shown in Figure~\ref{fig:local-and-global-gat-hops-qasper}. We can see an improvement in the performance from 0 to 4 (in the local graph network) and from 0 to 1 (in the global graph network). The experimental results demonstrate the importance of multi-hop attention among local and global nodes in the graph.

\subsection{Effects of the Global Graph Network}
\begin{figure}[t]
\centering
\includegraphics[width=\linewidth]{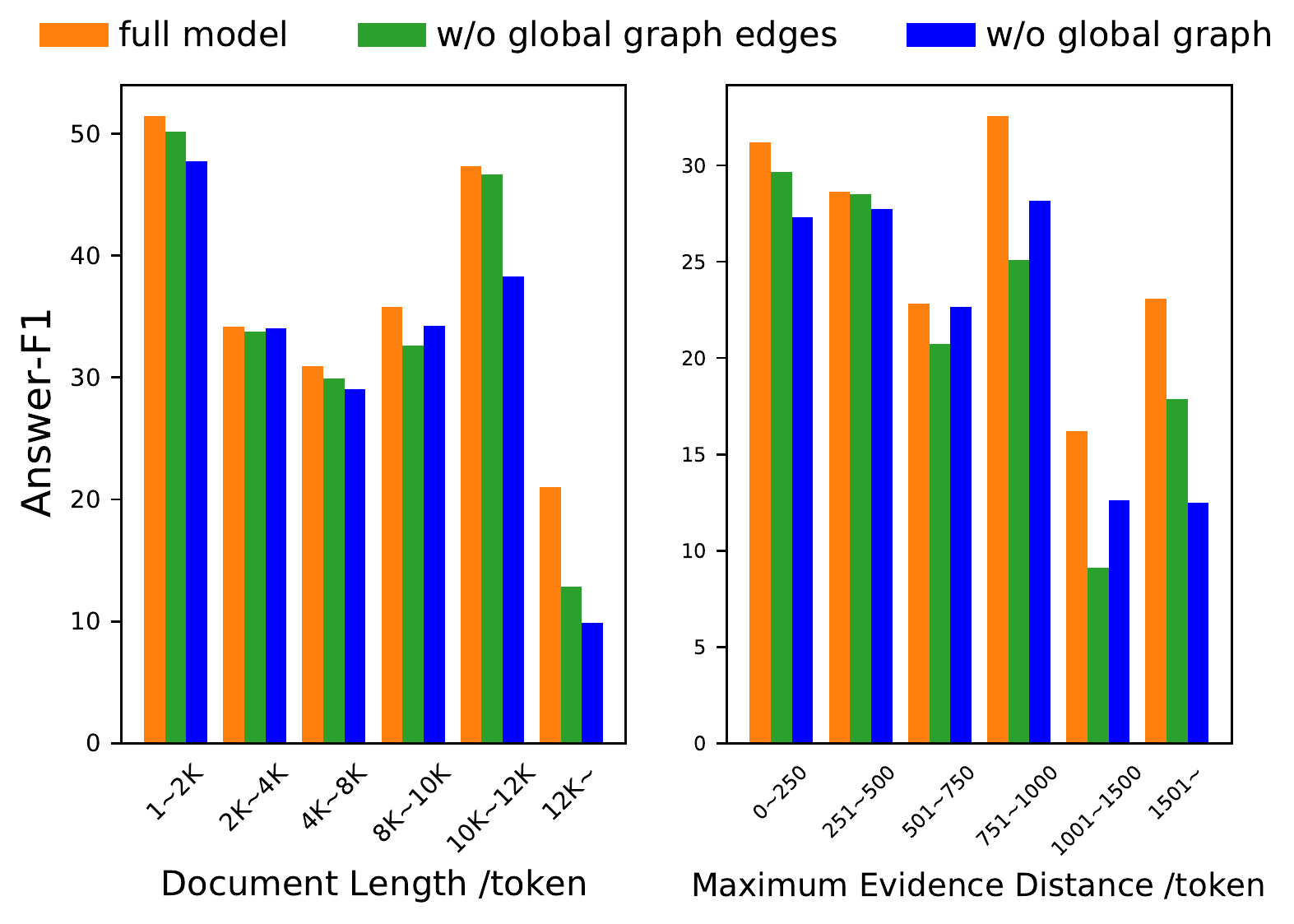}
\caption{\label{fig:ablation-on-evidence-span}Answer F1 of the proposed CGSN model and two ablated models on the dev set of Qasper. In the left subfigure, we evaluate each model by the context document length in each data instance. On the right side, the dev set is partitioned via the number of tokens between the first and the last gold evidence paragraphs in the context  document of each QA pair, called ``Maximum Evidence Distance''.}
\end{figure}

To better understand the functionality of the global graph, we conduct an experiment as shown in Figure~\ref{fig:ablation-on-evidence-span}. First of all, it illustrates that CGSN outperforms any other ablated versions in each range of ``document length'' or ``maximum evidence distance''. Besides, when the document length and the maximum evidence distance are large, the gap between CGSN and other versions becomes bigger, which demonstrates the ability of the global graph to capture long-range dependency. Furthermore, comparing the two ablated versions, it suggests that the edges of the global graph can enhance the ability of the model on long text understanding.

\subsection{Effects of Evidence Memory}

Redundancy is an essential problem in LDQA since in the answer generation phase, repetitious paragraphs could make the QA model confused (Appendix~\ref{appendix:case_study}). Therefore, it's crucial to select important and diverse evidence pieces for the QA model. To this end, we explore the effects of the evidence history module on redundancy reduction.

As shown in Table~\ref{tab:redundancy}, we evaluate the redundancy of the proposed model CGSN and the model ``w/o Mem''. $\textbf{REP}$ \citep{zhong2019searching} is used to calculate the repetitions in the generated summary. Based on it, we introduce $\textbf{REP}_{\text{inter}}$ to evaluate the redundancy among different paragraphs selected:
\noindent
\begin{equation}
    \text{REP}_{\text{inter}}=\frac{\text{CountUniq}\left(\{\text{ngram}_i\}_{i=1}^N\right)}{\sum_{i=1}^N\text{Count}(\text{ngram}_i)}
\end{equation}
\noindent
where $N$ is the number of selected paragraphs, $\text{Count}(\text{ngram}_i)$ is the number of n-grams in the paragraph $i$, $\text{CountUniq}\left(\{\text{ngram}_i\}_{i=1}^N\right)$
is the number of grams in all the selected paragraphs.

According to Table~\ref{tab:redundancy}, gold evidence obtains the least redundancy in them. Besides, it is found that without the evidence memory module, the redundancy of the proposed model gains. It demonstrates that the evidence memory can alleviate the redundancy problem in the evidence selection phase.

\begin{table}[t]
\fontsize{10}{11}\selectfont
  \centering
  \begin{tabular}{lccc}
    \toprule
    Models& $\textbf{REP}_{\text{inter}}\uparrow$ & $\textbf{REP}\uparrow$ & $\textbf{Score}_{\text{red}}\downarrow$ \\
    \midrule
    CGSN &  \textbf{0.8751} & \textbf{0.5708} & \textbf{-0.3166} \\
    w/o Mem &  0.8688 & 0.5661 & -0.3133 \\
    \midrule
    Gold &  0.9345 & 0.6301 & -0.5092 \\
    \bottomrule
  \end{tabular}
  \caption{Comparison between the full CGSN model and the model without evidence memory (w/o Mem). $\textbf{REP}$ \citep{zhong2019searching}, $\textbf{Score}_{\text{red}}$ \citep{chen2021training} and the proposed $\textbf{REP}_{\text{inter}}$ are used to evaluate the redundancy among the selected paragraphs. ``$\uparrow$'' denotes higher is better while ``$\downarrow$'' means lower is better. ``Mem'' denotes the evidence memory. ``Gold'' denotes the gold evidence.}
  \label{tab:redundancy}
\end{table}

\section{Conclusion}
To solve the problem of lacking global structure in LDQA methods, we propose Compress Graph Selector Network to capture the global structure over the long document when selecting evidence pieces. Extensive experiments demonstrate the strong performance of the model.

\section*{Limitations}

Since the proposed model is applied in an iterative way to read the long document, the limitation is the time cost. To alleviate the problem, we pack a series of paragraphs into a segment and encode them in parallel. However, methods better at parallelized computing could further save the required time for training and inference.

\section*{Acknowledgements}
The work is supported by National Key R\&D Plan (No. 2020AAA0106600), National Natural Science Foundation of China (No. U21B2009, 62172039 and L1924068).
We thank the EMNLP reviewers for their helpful feedback. We would like to acknowledge Yuming Shang for the helpful discussions.

\bibliography{anthology,custom}
\bibliographystyle{acl_natbib}

\appendix

\section*{Appendix}
\section{Algorithm}
We describe the detailed architecture of CGSN in Algorithm~\ref{algorithm:model}.
\label{sec:algorithm}
\begin{algorithm}[t]
\fontsize{11.2}{12}\selectfont
  \SetAlgoLined
  \KwData{LDQA training dataset $\mathcal{S}$, CGSN model $\mathcal{M}$, composed of a contextual \textbf{Encoder}, a Local Graph Network \textbf{LGN}, a Global Graph Network \textbf{GGN}}
  \KwResult{A fine-tuned model $\mathcal{M}'$.}
  Initialize the parameters of $\mathcal{M}$\;
  Set the evidence memory \textbf{Mem}$\leftarrow$\textit{None}\;
  Shuffle the dataset $\mathcal{S}\leftarrow\{(d^{(l)},q^{(l)},E^{(l)},a^{(l)})\}_{l=1}^S$;

  \ForEach{\textup{document }$d^{(l)}$ \textup{in} $\mathcal{S}$}{
      Split $d^{(l)}$ into segments $[P_1, P_2, ...,P_N]$, where $P_l=[p_{l_1}, ...,p_{l_N}]$\;

      \ForEach{\textup{segment }$P_l$ \textup{in} $d^{(l)}$}{
          Time step T$\leftarrow$ l\;
          \If{$T\neq$0}{
            write\_evidence(\textbf{Mem}, \textbf{GGN})\;
          }
          Segment $\mathbf{B}\leftarrow\{(q^{(l)}, p_{l_r})\}_{m=1}^{l_N}$\;
          \textbf{H} $\leftarrow$ \textbf{Encoder}($\mathbf{B}$)\;
          
          local\_graph\_init(\textbf{H},\textbf{LGN})\;
          
          \For{$l\leftarrow 1$ \KwTo \textup{Local\_hops}}{
          unidirect\_hops(\textbf{LGN})\;
          }
          
          multihead\_attention(\textbf{LGN},\textbf{GGN})\;
          
          \For{$l\leftarrow 1$ \KwTo \textup{Global\_hops}}{
          bidirect\_hops(\textbf{GGN})\;
          } 
          
          multihead\_attention(\textbf{GGN},\textbf{LGN})\;  
          
          p\_logits, loss $\leftarrow$ cal\_loss(\textbf{LGN}, $\mathbf{E}_{\text{P}_i}$)\;
          
          \textbf{Mem} $\leftarrow$ weighted\_sum(p\_logits, \textbf{LGN})\;
          update\_params($\mathcal{M}$,loss)\;
      }
  }
 
  \caption{\label{algorithm:model}Compressive Graph Selector Network}
\end{algorithm}
\begin{table*}[t]
\fontsize{11.3}{12}\selectfont
  \centering
  \begin{tabular}{p{.95\textwidth}}
    \toprule
    The proposed full model\\
    \midrule
    \textbf{Selected Paragraphs}:\\
    ......LiLi uses a Knowledge Store (KS) for knowledge retention. KS has four components: (i) \textcolor{blue}{Knowledge Graph} ( INLINEFORM0 ) ...... (ii) \textcolor{blue}{Relation-Entity Matrix} ( INLINEFORM2 ) ...... (iii) \textcolor{blue}{Task Experience Store} ( INLINEFORM15 ) ...... (iv) \textcolor{blue}{Incomplete Feature DB} ( INLINEFORM29 ) ...... LiLi also uses \textcolor{red}{a stack, called Inference Stack} ( INLINEFORM0 ) to hold query ......\\
    \textbf{Generated Answer}: \textcolor{blue}{Knowledge Graph, Relation-Entity Matrix, Task Experience Store, Incomplete Feature DB}, \textcolor{red}{Stack}\\
    \midrule
    Ablated Model (w/o global graph)\\
    \midrule
    \textbf{Selected Paragraphs}:    ......LiLi uses a Knowledge Store (KS) for knowledge retention. KS has four components: (i) \textcolor{blue}{Knowledge Graph} ( INLINEFORM0 ) ...... (ii) \textcolor{blue}{Relation-Entity Matrix} ( INLINEFORM2 ) ...... (iii) \textcolor{blue}{Task Experience Store} ( INLINEFORM15 ) ...... (iv) \textcolor{blue}{Incomplete Feature DB} ( INLINEFORM29 ) ...... \\
    \textbf{Generated Answer}:\\
    \textcolor{blue}{Knowledge Graph, Relation-Entity Matrix, Task Experience Store, Incomplete Feature DB}\\
    \bottomrule
  \end{tabular}
  \caption{Comparison between the proposed full model and the model ablated global graph. In this case, the length of the document is 8768, which cannot be processed by most of the pre-trained models in one step. \textcolor{blue}{Matched content} is colored in blue while  \textcolor{red}{mismatch content} is colored in red.}
  \label{tab:case1}
\end{table*}
\begin{table*}[t]
\fontsize{11.3}{12}\selectfont
  \centering
  \begin{tabular}{p{.95\textwidth}}
    \toprule
    The proposed full model\\
    \midrule
    \textbf{Selected Paragraphs}:\\
    Two datasets are exploited in this article ....... \textcolor{blue}{NUBes BIBREF4} is a corpus of around 7,000 real medical reports ...... the second set of experiments uses the \textcolor{blue}{MEDDOCAN 2019 shared task competition dataset} ....... \\
    \textbf{Generated Answer}: NUBes BIBREF4, MEDDOCAN 2019 shared task competition dataset\\
    \midrule
    Ablated Model (w/o evidence memory)\\
    \midrule
    \textbf{Selected Paragraphs}:    Two datasets are exploited in this article ....... \textcolor{blue}{NUBes BIBREF4} is a corpus of around 7,000 real medical reports ...... the second set of experiments uses the \textcolor{red}{MEDDOCAN 2019 shared task competition dataset} .......
 		We have compared this BERT-based sequence labelling against other methods and systems. One of the experiments uses the \textcolor{red}{MEDDOCAN 2019 shared task dataset}......\\
    \textbf{Generated Answer}: NUBes BIBREF4, MEDDOCAN 2019 shared task competition dataset\\
    \textbf{Generated Answer}:\\
    NUBes BIBREF4, \textcolor{red}{BERT-based sequence labelling approach}\\
    \bottomrule
  \end{tabular}
  \caption{Comparison between the proposed full model and the model without evidence memory. \textcolor{blue}{Matched content} is colored in blue while \textcolor{red}{redundant content} is colored in red.}
  \label{tab:case2}
\end{table*}

\section{Analysis of Memory Usage}
In this section, we analyze the memory usage of the end-to-end method LED and the proposed CGSN. Assuming the length of a document is $L\ (L\geq 4K)$, the local window size\footnote{Attention to the $\frac{W}{2}$ tokens ahead and $\frac{W}{2}$ tokens behind.} is $W (W\geq 512)$, the number of global tokens is $G_t$, the memory usage of LED method is $O\left(L(W+G_t)\right)$. When $G_t\ll W$, the memory usage is $O\left(LW\right)$. For CGSN, set the paragraph number in a segment as $B$. For a fair comparison, the maximum length of a paragraph is $W_{\text{half}}=\lfloor \frac{W}{2}\rfloor$. Therefore, the memory usage in CGSN is $O\left(BW^2_{\text{half}}+f_{\text{local}}(B,W)+M_{\text{global}}\right)$, where $f_{\text{local}}$ is the function of memory usage in local graph and $M_{\text{global}}$ is the memory usage of global graph, which is pre-defined as a constant. It means that memory usage of CGSN is constant to the document length while LED is nearly linear to the length (when $W$ becomes larger, the linearity will convert into quadratic).

\section{Answer Generator}
\label{appendix:answer_generator}
Following the `led\_base\_smaller\_context' setting\footnote{https://github.com/allenai/qasper-led-baseline} in \citet{dasigi2021dataset}, a vanilla LED is used as the answer generator, which is trained with gold evidence coupled with QA pairs in the training set. To evaluate the performance of an evidence selection method in perspective of answer generation, for each question and its long document. The evidence selection method is firstly used to provide evidence pieces given the question and the long document. And then, the corresponding selected evidence and the question are concatenated and fed into the LED model. After that, the LED model generates answer tokens for the question. The generated answer tokens are then used to calculate the `Answer-F1' score given the gold answer.

\section{Case Study}
\label{appendix:case_study}

we conduct the case study in this section to investigate the effectiveness of the global graph and the evidence memory qualitatively.
As shown in Table~\ref{tab:case1}, the proposed model can correctly select all the evidence pieces across different positions of the text and integrate them into the complete evidence.\footnote{The original case in the dataset has some errors and we correct them in Table~\ref{tab:case1}.} However, for the model without the global graph, a piece of evidence is lost. It indicates that in a long document, the understanding of long-range is crucial for evidence selection and answer generation.

Table~\ref{tab:case2} compares the proposed model and the one without evidence memory, where the ablated model chooses ``MEDDOCAN 2019 shared task'' twice as well as adds some other noise information (``BERT-based sequence labeling approach'') which makes the model confused in the answer generation phase.

\end{document}